\documentclass{article}

\usepackage{PRIMEarxiv}

\usepackage[utf8]{inputenc} 
\usepackage[T1]{fontenc}    
\usepackage{hyperref}       
\usepackage{url}            
\usepackage{booktabs}       
\usepackage{amsfonts}       
\usepackage{nicefrac}       
\usepackage{microtype}      
\usepackage{lipsum}
\usepackage{fancyhdr}       
\usepackage{graphicx}
\usepackage{multirow}
\graphicspath{{media/}}     
\usepackage{amsmath}
\title{Quantifying the uncertainty of model-based synthetic image quality metrics
}

\author{
  Ciaran Bench, Spencer Thomas \\
  Department of Data Science and AI \\
  National Physical Laboratory \\
  Hampton Road, Teddington,TW11 0LW  UK\\
  \texttt{\{ciaran.bench, spencer.thomas\}@npl.co.uk} \\
}

\begin{document}
\maketitle

\begin{abstract}
The quality of synthetically generated images (e.g. those produced by diffusion models) are often evaluated using information about image contents encoded by pretrained auxiliary models. For example, the Fr\'{e}chet Inception Distance (FID) uses embeddings from an InceptionV3 model pretrained to classify ImageNet. The effectiveness of this feature embedding model has considerable impact on the trustworthiness of the calculated metric (affecting its suitability in several domains, including medical imaging). Here, uncertainty quantification (UQ) is used to provide a heuristic measure of the trustworthiness of the feature embedding model and an FID-like metric called the Fr\'{e}chet Autoencoder Distance (FAED). We apply Monte Carlo dropout to a feature embedding model (convolutional autoencoder) to model the uncertainty in its embeddings. The distribution of embeddings for each input are then used to compute a distribution of FAED values. We express uncertainty as the predictive variance of the embeddings as well as the standard deviation of the computed FAED values. We find that their magnitude correlates with the extent to which the inputs are out-of-distribution to the model's training data, providing some validation of its ability to assess the trustworthiness of the FAED. 
\end{abstract}


\section{Introduction}

Generative modelling is used in a wide range of applications; e.g to augment training datasets \cite{tanaka2019data} or anomaly detection \cite{di2019survey}. While there are several variants of the task, in the most generic case the aim is to learn to randomly sample from the data distribution $P(y)$ expressed as a discrete set of images $Y$, where  $y_1, y_2, ... , y_N \in Y$ \cite{creswell2018generative}. Several different generative modelling frameworks have been proposed, but the two most commonly implemented are generative adversarial networks \cite{gui2021review} and diffusion models \cite{yang2023diffusion}. Both types of model are challenging to train, and are prone to producing artefacts and hallucinations. This motivates the need to carefully validate their performance \cite{betzalel2022study,ahmed2025style}. Validation is challenging for several reasons, including the absence of a ground truth data distribution and the difficulty of defining quantitative measures of characteristic structural and stylistic content \cite{salehi2020generative,borji2022pros}. Instead, it is usual to have a discrete approximation of it defined by the training set (e.g. $Y$ as described above). 

The optimal procedure for evaluating model performance depends on the downstream application. For image-to-image translation models, it is important to assess whether the structural content within the input is present in the translated image. However, this is not necessarily the case for generic image generation tasks. When generated images are used to supplement training data, the performance of the model trained on this data can provide an indication of quality \cite{wu2025pragmatic}, though this is computationally expensive, and applies to a narrow use-case. For most applications of generative modelling, assessing whether the style of the generated images (i.e. the feature representations for the objects/structures in an image, as well as its semantic content) is in line with the distribution of real images is of critical importance. Stylistic content is often assessed by comparing the distributions of characteristic features detected from the generated data, and data drawn from the ground truth distribution. These are typically detected using pre-trained classification networks and assessed using metrics such as the Fr\'{e}chet Inception Distance (FID) \cite{heusel2017gans}. Here, generated images $x_1, x_2, ..., x_N \in X$ and those from some real, reference distribution $y_1, y_2, ..., y_N\in Y$ (where N is the total number of images) are processed by an InceptionV3 network pretrained to classify ImageNet. The mean $\mu$ and covariance $\Sigma$ of the activations 
produced from the synthetic images are used to parameterise a multivariate Gaussian distribution $\mathcal{N}_{{x}}(\mu_{x}, \Sigma_{x})$, that is then compared to the corresponding distribution computed from the reference set of images $\mathcal{N}_{y}(\mu_y, \Sigma_y)$, in an unpaired manner, with a Fr\'{e}chet distance:
\begin{equation}
\label{eq:fd}
\begin{split}
        \mathrm{FID}(X,Y) =
        \left\|\mu_{{x}}-\mu_y\right\|_2^2+\text{tr}\left(\Sigma_{{x}}+\Sigma_y-2\left(\Sigma_{{x}} \Sigma_y\right)^{\frac{1}{2}}\right).
    \end{split}
\end{equation}

However, models (classifiers or otherwise) trained on natural images are not always effective at detecting the characteristic features of other datasets, making them unsuitable as feature extractors for these evaluation metrics. For instance, this is a well known problem in medical imaging where the FID is generally considered to be a suboptimal and inadequate image quality metric \cite{deshpande2025report,tronchin2021evaluating,wu2025pragmatic}. This is a fundamental limitation of any variants of the FID that utilise other feature extraction models, such as the Fr\'{e}chet Autoencoder Distance \cite{buzuti2023frechet}, and raises concerns about the trustworthiness of the metrics. 
Knowledge of how the uncertainty in the feature embedding affects the accuracy of the metric could be used to assess the calculated metric's trustworthiness and the suitability of the feature extraction model for a particular domain application.

In this work, we train a convolutional autoencoder (CAE) based feature extraction model with approximate variational learning (Monte Carlo dropout) to model the epistemic uncertainty in its embeddings. We demonstrate how the predictive distribution of latent representations can be used to estimate a corresponding predictive distribution of Fr\'{e}chet Autoencoder Distance (FAED) values (akin to the FID, but considers the latent representations produced by a pretrained CAE). We investigate whether the predictive variance in the embeddings, as well as the standard deviation in the corresponding distribution of FAED values, can indicate the effectiveness of the feature embedding model and provide a heuristic estimate of the uncertainty of the FAED.

\section{Methods}

\subsection{Monte Carlo dropout}
Bayesian optimisation is a commonly used framework for estimating epistemic  uncertainty (reducible model uncertainty) for predictive models \cite{jospin2022hands}. However, this is known to not scale well with the number of parameters used in deep learning models. Similarly, approximation schemes like variational inference also do not scale well with the number of parameters \cite{shen2024variational}. Monte Carlo dropout is an approximation of variational inference, where in the most generic implementation, one simply trains a model with dropout layers distributed liberally throughout the architecture, and leaves them active during evaluation \cite{gal2016dropout}. At test time, each test input is evaluated several times producing a distribution of predictions whose variance is typically interpreted as epistemic uncertainty (for this generic implementation). We refer to this evaluation procedure as dropout ensembling. We opt for Monte Carlo dropout instead of other scalable uncertainty quantification techniques such as deep ensembles given its ease of implementation, efficiency, and to avoid challenges with reparameterisation \cite{magers}.

\subsection{Training}
We train a model $G$ composed of an encoder $E$ and decoder $D$ modules, where $E$ learns to transform an input image $x_i$ into a compressed representation $l_{i}$. This is then used as input to the decoder that will ideally reconstruct the input image, i.e. $D(E(x_i))\approx x_i$. Dropout regularisation with a dropout rate of 10 \% is applied to each layer of the encoder. We use the pixel-wise mean squared error as the loss function, using the sum of all pixels to aggregate values. The model was trained on the ImageWoof dataset (\url{https://github.com/fastai/imagenette}) using the default split (9,035 training examples, and 3,929 validation examples). Each image was resized to 128 by 128 pixels and normalised to have a range of 0-1. We used Adam as the optimiser and a batch size of 16. The architecture was composed of three encoder convolutional layers with ReLU activations, a stride of two, and a kernel size of four with the following output channels: $128\rightarrow256\rightarrow512$. The final encoder layer is followed by a flatten layer and a linear layer with 256 output nodes (the length of the latent dimension). The decoder begins with a linear layer that takes the latent representation as an input and outputs a 1D array of length 131072 which is reshaped and then processed by three transposed convolutional layers with ReLU activations, a stride of two, and a kernel size of four, with the following output channels: $256\rightarrow128\rightarrow3$. The parameters associated with the smallest validation loss (over 25 total epochs) were used for evaluation. A small length of 256 elements for the latent representation was chosen to ensure stability in the calculation of the FAED metric.

\subsection{Evaluation}
We use dropout ensembling at evaluation to produce a distribution of 200 latent representations for each test input. Specifically, a pretrained CAE is used to extract latent representations $l_{i,j,k}$ (i.e the activations from the output of the bottleneck layer) from a set of images $\hat{x}_1, \hat{x}_2, ...,\hat{x}_i,... \hat{x}_N \in \hat{X}$, where $j$ represents the index of the latent representations produced for a given input $i$ via dropout ensembling, and $k$ is each element of a given latent representation. One expression of uncertainty over the whole test set is given by the mean over the element-wise variance of the sampled embeddings:
\begin{equation}
    \text{pVar} = \mathbb{E}_{k} \left[ \mathbb{E}_{i} \left[ \text{Var}_{j} \left( l_{i,j,k} \right) \right] \right]. 
\end{equation}
The variance in the embeddings can provide an indication of model's capability to accurately detect its contents.

An alternative expression of uncertainty that comes from the standard deviation ($\sigma$) of the FAED scores over $j$ is denoted by $\sigma_{\text{FAED}}$.  
The mean $\mu^{j}_{\hat{x}}$ and covariance $\Sigma^{j}_{\hat{x}}$ of the activations $l_{1,j},l_{2,j},...,l_{i,j},...l_{N,j}$ are used to parameterise a multivariate Gaussian distribution $\mathcal{N}_{{\hat{x}}}(\mu^{j}_{\hat{x}}, \Sigma^{j}_{\hat{x}})$, that is then compared to the corresponding distribution computed from a separate set of images (in an unpaired manner) $\mathcal{N}_{y}(\mu_y, \Sigma_y)$ with a Fr\'{e}chet distance:
\begin{equation}
\label{eq:fd}
\begin{split}
        \mathrm{FAED}^{j}(\hat{X},Y) = d_F\left(\mathcal{N}_{{\hat{x}}}(\mu^{j}_{\hat{x}}, \Sigma^{j}_{\hat{x}}), \mathcal{N}_{y}\left(\mu_y, \Sigma_y\right)\right)^2= \\
        \left\|\mu^{j}_{\hat{x}}-\mu_y\right\|_2^2+\text{tr}\left(\Sigma^{j}_{\hat{x}}+\Sigma_y-2\left(\Sigma^{j}_{\hat{x}} \Sigma_y\right)^{\frac{1}{2}}\right).
    \end{split}
\end{equation}

\begin{equation}
\label{eq:vfd}
\begin{split}
        \mathrm{\sigma_{\text{FAED}}}(\hat{X},Y) = \sigma_{j} \left( \mathrm{FAED}^{j}(\hat{X},Y) \right)
    \end{split}
\end{equation}
In this case we compute 200 FAED scores, where the $j$th score is computed using the $j$th sampled latent representations from each test input (see Equation \ref{eq:fd}).

We evaluate our model with various image datasets to assess whether the FAED and $\sigma_{\text{FAED}}$ scores are higher for images that are increasingly out-of-distribution relative to the CAE's training set. As a baseline, we evaluate the CAE's validation set (ImageWoof), where $\hat{X}$ is given by the first half, and $Y$ by the second half. The embeddings from the second half of ImageWoof's validation set was used as $Y$ for all other evaluations of the $\sigma_{\text{FAED}}$. In order of increasing domain gap relative to the CAE's training set (ImageWoof), we also evaluate on: ImageWoof validation set augmented with noise (drawn from a Gaussian with mean zero and standard deviation of 2~\% of the max value of the image), ImageWoof validation set augmented such that five mini (31x31 pixels) randomly rotated versions of the input are overlaid in random locations on each input image, ImageWoof validation set augmented using the same approach but with randomly rotated natural Imagenette images (not including the English Springer dog class) (\url{https://github.com/fastai/imagenette}), and the Imagenette dataset with the English Springer class removed. Examples of these augmentations are given in Table ~\ref{tab:fvead_comparison}.

\subsection{Related work}
\subsubsection{Learning stochastic embeddings for comparing data}
Several works discuss the concept of learning stochastic embeddings and how the variance in the embeddings acquired for a given input can be used to infer whether the input image shares similar properties with the model's training data (i.e whether its position in the latent manifold defined by the encoded training data has low uncertainty) \cite{catoni2024uncertainty,shi2019probabilistic,an2023maximum,butepage2021gaussian}. While this may indirectly encode the model's ability to detect image contents, these approaches do not explicitly model epistemic uncertainty. In \cite{catoni2024uncertainty}, the authors assess the inherent uncertainty in the latent representations learned by a VAE (`inherent' due to the probablistic optimisation), indicating that expressions of uncertainty given by the aggregated predicted means and variances can be used to perform out-of-distribution detection (relative to the model's training set), and detect variations in contrast/degrees of image corruption. Other work \cite{shi2019probabilistic} uses the variances predicted by a VAE as a measure of uncertainty in the embeddings of face images which may have significantly different properties than the training data. These uncertainties are then used to improve facial recognition strategies (e.g. focus on regions of the latent representation with lower uncertainty). Instead of using the generic VAE framework where a Gaussian prior is imposed on the predicted distribution of latent representations, \cite{an2023maximum} instead proposed an entropy-based approach as a measure of uncertainty. This is optimised with the underlying assumption that the most optimal distribution is the one with the largest entropy (with the inherent tradeoff with the expressivity of the latent representations). They found that this helped improve model robustness, and that the uncertainties provided an effective confidence measure, facilitating risk-controlled person identification. In \cite{an2015variational}, an autoencoder was used to learn anomaly detection. For each input, they sample several latent representations, each of which were used as inputs to the decoder, which outputs the mean and variance of a normal distribution. The likelihood of the original input under each normal distribution output by the decoder were calculated and averaged together, where low values correspond to anomalies. They found that this approach outperformed more generic autoencoder-based anomaly detection algorithms. 

The studies mentioned above establish a precedent for using the variance in stochastic embeddings to assess whether a given input has similar qualities to a model's training set. In principle, this could be used to derive useful heuristic uncertainty estimates for model-based image quality metrics, as it indirectly encodes some information about how well the model may detect an image's contents. With that said, they do not explicitly consider epistemic uncertainty. This more directly encodes the extent to which the feature extraction model is capable of detecting information characterising image contents. It is critical to consider this source of uncertainty when deriving heuristic measures of uncertainty for model-based image quality metrics, as they are fundamentally limited by the performance of the feature-embedding model.

\subsection{Uncertainty quantification for autoencoder models}
There is precedent for the use of uncertainty quantification (including Monte Carlo dropout for epistemic uncertainty) on autoencoders, including VAEs \cite{miok2019generating,magers,warburg2023bayesian}. However, this is not necessarily performed to provide some direct measure of the uncertainty in the embeddings themselves; e.g. using a Gaussian Negative Log-Likelihood loss to estimate aleatoric uncertainty in image reconstructions, and using anchored ensembling to estimate epistemic uncertainty \cite{yong2020bayesian}. They show that the estimated uncertainties correlate with expectations about the performance of the model with data that has been augmented to various extents. 

Monte Carlo dropout has been implemented on VAE and CAE models by adding dropout regularisation to the decoder layers, and demonstrated performance benefits when applying this regularisation to the models \cite{miok2019generating} (though the authors do not use estimated uncertainties directly).
Deep ensembles can be used to quantify the uncertainties in the embeddings of models, which encodes epistemic uncertainty \cite{magers} . They discuss techniques to remedy the reparamaterisation problem (i.e. the fact that the latent representations learned by each model may not be comparable), which generally makes the use of ensemble-based UQ unappealing for applications that compare latent representations. 

The authors in \cite{warburg2023bayesian} constructed a model with a variational bottleneck layer akin to a VAE, but used it to learn a classification task. The stochastic nature of the information bottleneck naturally lends itself to UQ (e.g. a distribution of predictions can be acquired by sampling several bottleneck representations for a given input). The uncertainties were found to exhibit some degree of calibration. Though, here, epistemic uncertainty is not considered in their experiments.

\subsection{Autoencoder variants of the FID}
Comparisons between the autoencoder-derived latent representations produced from  populations of images has been used to compare the similarity of datasets \cite{buzuti2023frechet,fan2025enhancing}. Buzuti et al \cite{buzuti2023frechet} compared the activations of the bottleneck layer of a Vector Quantised Variational Autoencoder acquired from two populations of images with a Fr\'{e}chet distance (akin to the Fr\'{e}chet Inception Distance). They found that their distance metric correlated with expected differences in image characteristics with varying augmentation strengths. In \cite{fan2025enhancing}, a similar analysis was performed with a generic CAE. In, \cite{denouden2018improving} they instead used a Mahalanobis distance for out of distribution detection. Inspired by \cite{buzuti2023frechet}, we employ a CAE-based model for our experiments. This enables us to efficiently train a domain-specific model for our experiments.

\subsection{Our contributions}
Model-based image quality metrics like the FID require that the feature embedding model can effectively encode image contents. While the trustworthiness of these metrics depends on the epistemic uncertainty in the embeddings (among other factors, such as the stability of the covariance calculation involved with the FID), this is not typically considered when implemented in practice. Here, we use Monte Carlo dropout to model the epistemic uncertainty of the embeddings, and acquire a distribution of embeddings for each input image. We then derive heuristic uncertainty estimates of the FAED by computing the predictive variance of these embeddings, as well as the standard deviation of the corresponding distribution of FAED values. We provide some validation that our heuristic estimates of uncertainty correlate with the extent to which the inputs are out-of-distribution of the model's training data. This provides some indication that they may be used to assess the trustworthiness of the feature embedding model's ability to detect image contents, and hence, the trustworthiness of the FAED. This demonstrates how UQ may be used to assess the trustworthiness of related model-based image quality metrics.

\begin{table}[htbp]
  \centering
  \begin{tabular}{|c|c|c|c|}
    \hline
    \textbf{Test Data (with example input)} & \textbf{Mean FAED} & \textbf{$\sigma_{\text{FAED}}$} & \textbf{pVar}\\
    \hline
    \parbox[c]{5cm}{\centering ImageWoof (baseline) \\ 
     \begin{minipage}{0.2\textwidth}
        \centering
        \includegraphics[width=\linewidth]{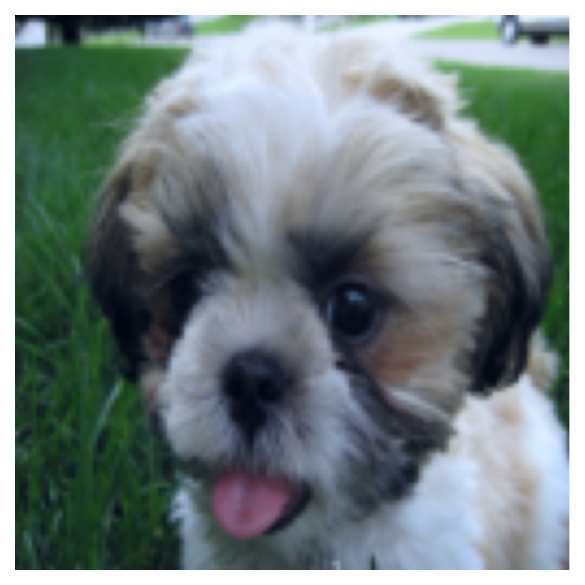}
      \end{minipage}} & 25.923 & 0.019& 0.0051\\
    \hline
    \parbox[c]{5cm}{\centering ImageWoof + 2 \% noise \\ 
     \begin{minipage}{0.2\textwidth}
        \centering
        \includegraphics[width=\linewidth]{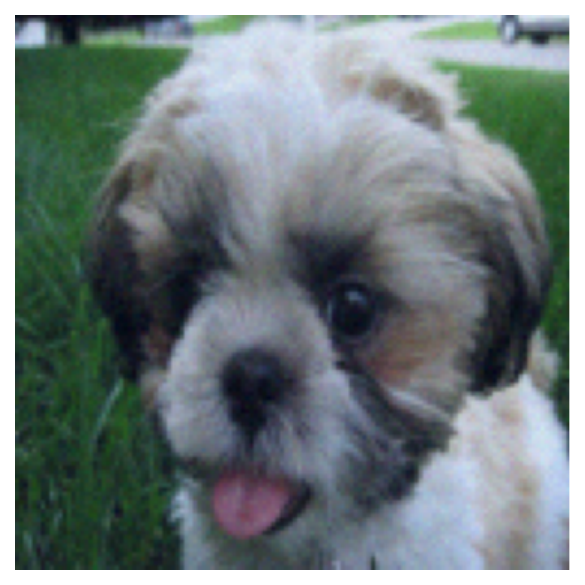}
      \end{minipage}} & 29.183 & 0.021&0.0042 \\
    \hline
    \parbox[c]{5cm}{\centering ImageWoof + five inputs overlaid\\ 
     \begin{minipage}{0.2\textwidth}
        \centering
        \includegraphics[width=\linewidth]{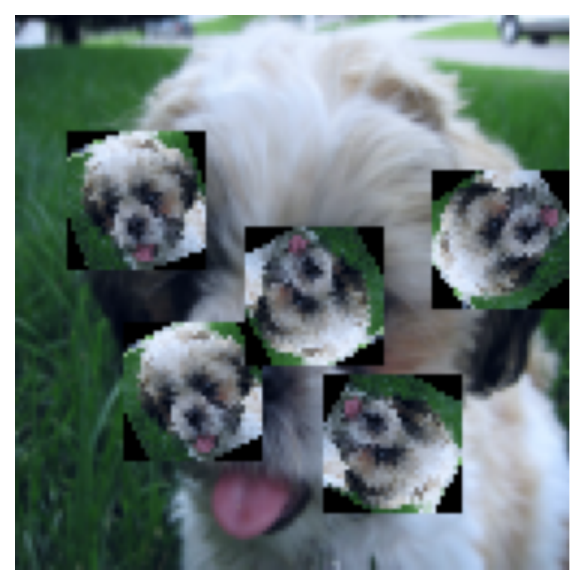}
      \end{minipage}} & 50.276 & 0.030&0.0062 \\
    \hline
    \parbox[c]{5cm}{\centering ImageWoof + five random Imagenette images (no dog) overlaid \\ 
     \begin{minipage}{0.2\textwidth}
        \centering
        \includegraphics[width=\linewidth]{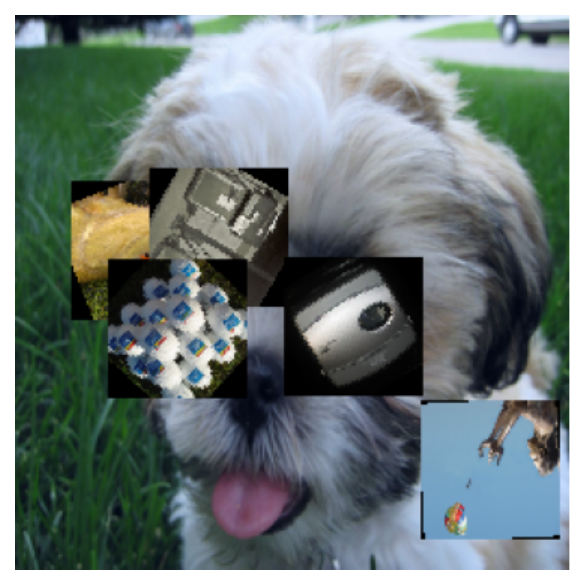}
      \end{minipage}} & 62.225 & 0.042& 0.0070 \\
    \hline
    \parbox[c]{5cm}{\centering Imagenette (no dog) \\ 
     \begin{minipage}{0.2\textwidth}
        \centering
        \includegraphics[width=\linewidth]{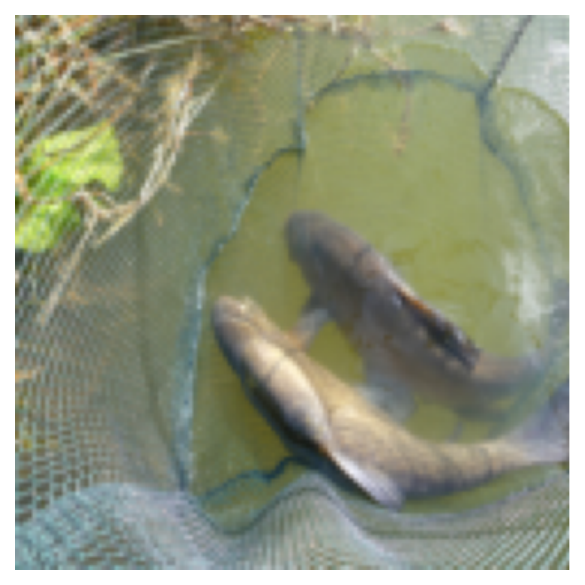}
      \end{minipage}} & 143.932 & 0.085&0.0082 \\
    \hline
  \end{tabular}
  \caption{Example inputs, and the mean FAED, $\sigma_{\text{FAED}}$, and pVar for each dataset. FAED and $\sigma_{\text{FAED}}$ scores compare the test data with the same reference dataset composed of the second half of the ImageWoof validation set. ImageWoof test inputs are augemented versions taken from the first half of the validation set.}
  \label{tab:fvead_comparison}
\end{table}
\begin{figure}[h!]
    \centering
    \includegraphics[width=0.49\linewidth]{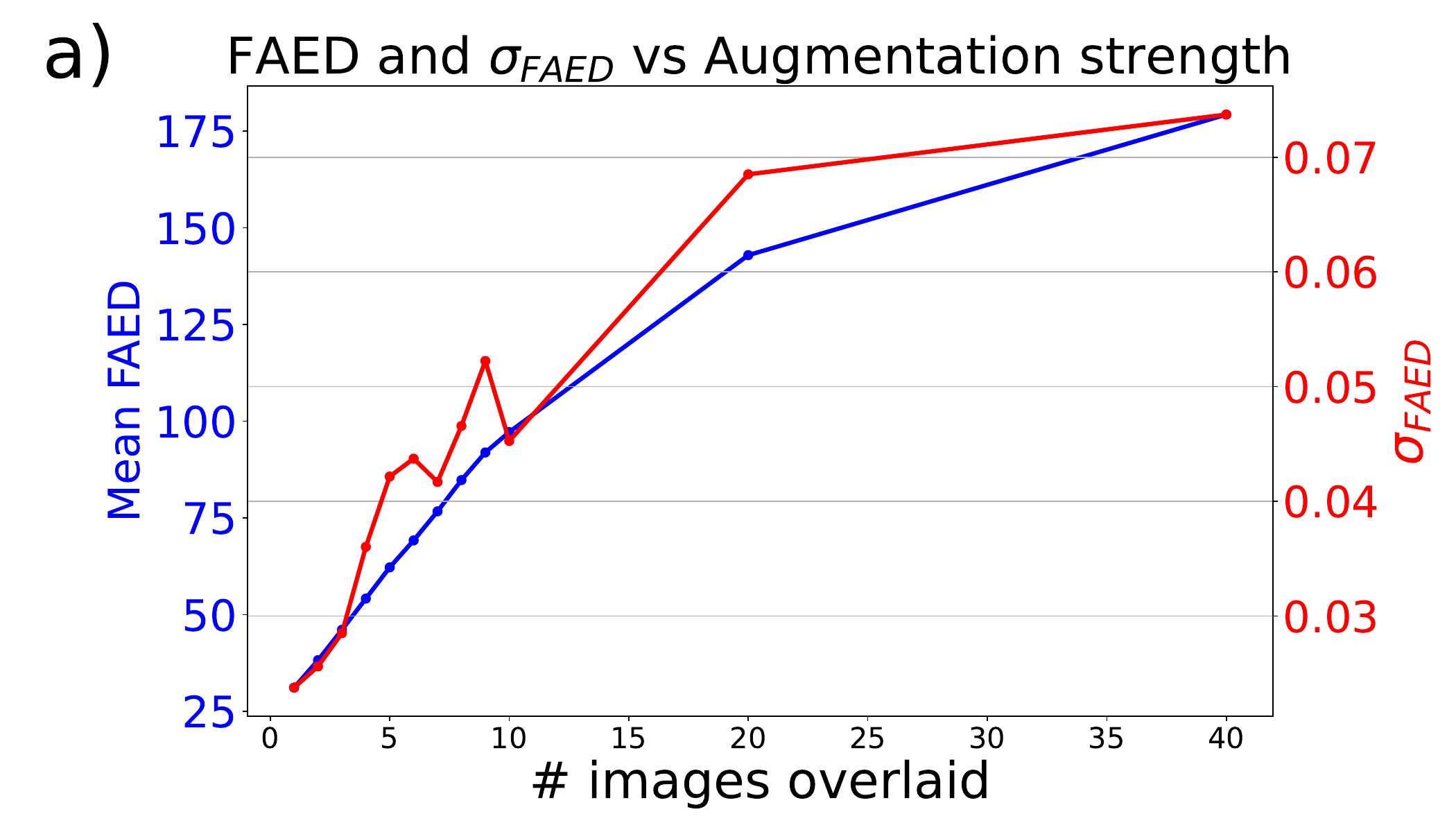} 
    \includegraphics[width=0.49\linewidth]{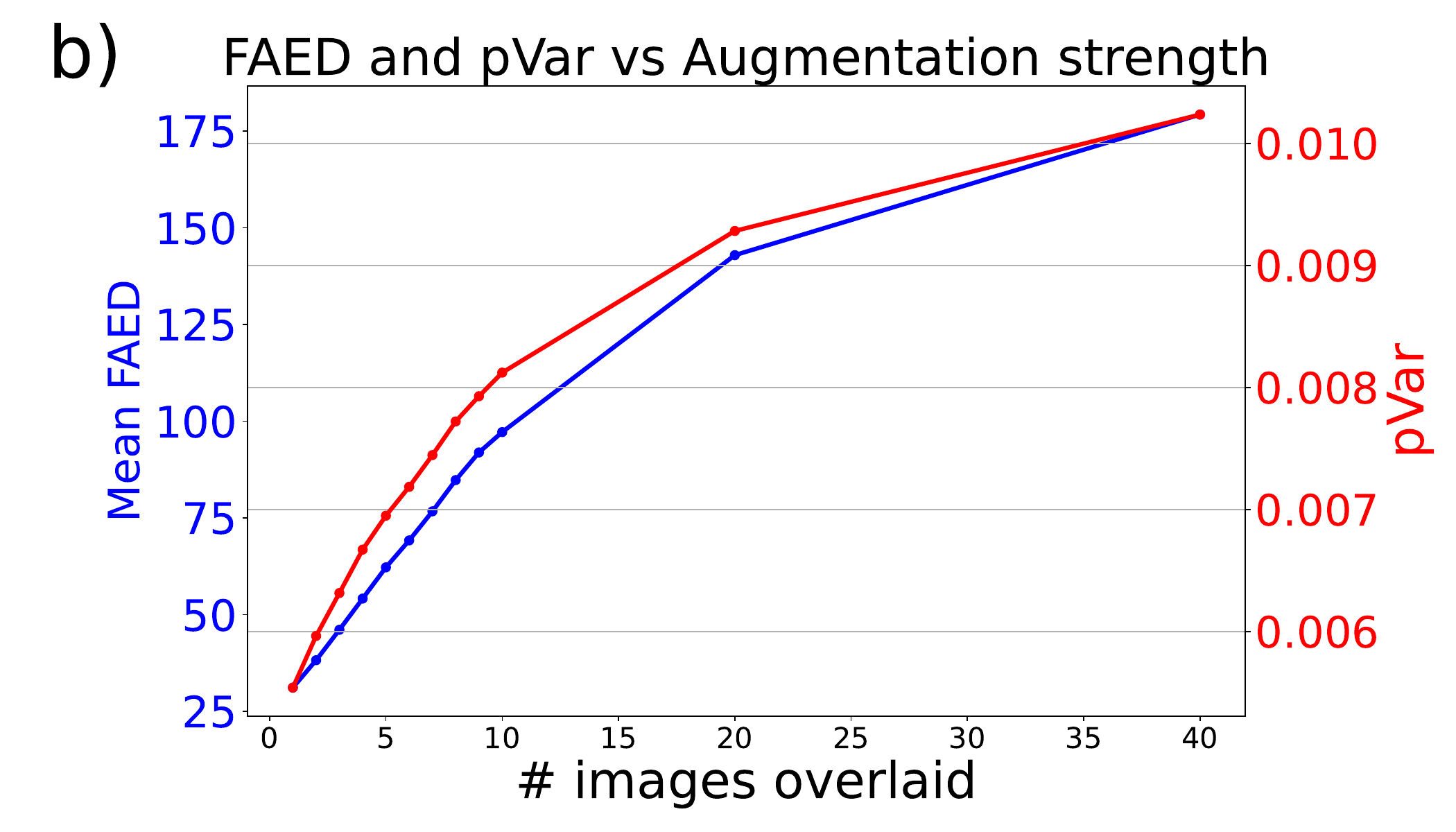} 
    \includegraphics[width=0.8\linewidth]{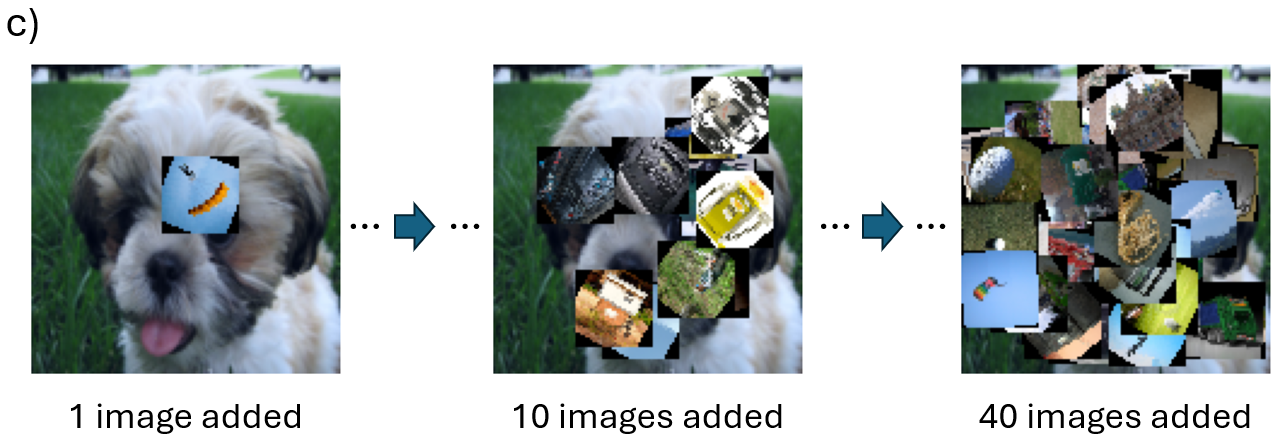}
    \caption{a) FAED and $\sigma_{\text{FAED}}$ vs the number of randomly rotated mini Imagenette images (no English Springer images) added to the ImageWoof inputs. b) the same plot but for FAED and pVar. We see all three metrics correlate strongly with the extent of the augmentation and therefore, the extent to which the inputs are out-of-distribution relative to the training data. c) example inputs for three augmentation cases.}
    \label{fig:noise_exp}
\end{figure}

\section{Results}
We find that the FAED scores themselves correlate with the extent to which the inputs are out of distribution from the model's training data. That is,  the highest scores are observed for the natural Imagenette images, the second highest for ImageWoof's validation set with natural images overlaid, the third highest was for ImageWoof validation set overlaid with miniature versions of themselves, and the fourth highest was for the  ImageWoof validation images augmented with 2 \% noise. 

We find similar trends for $\sigma_{\text{FAED}}$ and pVar, though the noise-augmented ImageWoof images were found to have a smaller pVar value than the unaugmented ImageWoof images. Unlike the other forms of data augmentation considered, noise augmentation reduces the complexity of input features by smoothing out finer details. The deeper layers of a model encoding images \textit{without} noise augmentation may encode highly specific and complex features. Here, the type of information that is detected may vary considerably if some neurons are dropped, given complex features are defined by the spatial arrangement of simpler features, and where each \textit{element} of a deep layer kernel may detect the presence of a simpler feature. This could make the variability in the encoding of more complex features more susceptible to dropout ensembling, increasing the variance in the latent representations. A model processing smoothed out images could have lower predictive variance as there is less complexity to detect, reducing the effect of the dropout on the final layers. Dropping connections in the filters of earlier layers could have less dramatic consequences on the information encoded compared to deeper layers given the lower receptive field. Further investigation of this behaviour will be the subject of future work. Nonetheless, we observe higher $\sigma_\text{FAED}$ for the noise augemented images relative to the unaugmented images. This suggests that while pVar may be a preferable way to capture the models's epistemic uncertainty in information encoding, this metric alone is not sufficient to capture how the covariability of FAED's terms contributes to the total variance of the metric. While smoothed out features may correspond with lower predictive variances, the covariance of the FAED terms may be smaller or larger in magnitude than for unaugmented images resulting in higher variability in the metric.

For the inputs augmented with Imagenette examples, we find that all three metrics scale with the strength of the augmentation (in this case, the number of Imagenette examples overlaid onto each input) as seen in Fig. \ref{fig:noise_exp}. This indicates that both metrics are sensitive to gradual changes in image content detected by the CAE. 

It is important to note that in our experiments, we consider only one version of the feature embedding model trained with a single dropout rate. However, it is well known that the dropout rate can influence the calibration of predicted uncertainties \cite{gal2016dropout}. Further investigations of how different parameterisations might affect the quality of the predicted uncertainties will be the subject of future work.

\section{Conclusion}
Model-based metrics for assessing the quality of synthetically generated images rely on the assumption that the auxiliary feature encoding model can accurately detect the contents/characteristic features of images. Here, we have shown how modelling the uncertainty in the embeddings can help provide a notion of the trustworthiness of the feature embedding model, and its corresponding image quality metric: the FAED.

We find that both the magnitude of the standard deviation ($\sigma_{\text{FAED}}$) of the distribution of FAED values we compute using the sampled embeddings, and the predictive variance of these embeddings (pVar) generally correlate with the extent to which the model inputs are out-of-distribution of its training data. However, we note that pVar decreased with noise augmentation, suggesting that augmentation techniques that smooth out/simplify feature representations may reduce uncertainties derived with dropout ensembling. Furthermore, this highlights how pVar does not capture how the correlations between the terms of the FAED influence its variability, highlighting the need to present both expressions of uncertainty for a comprehensive evaluation of trustworthiness. Broadly, our results suggests that $\sigma_{\text{FAED}}$ and pVar values can be used as a heuristic estimate of the uncertainty in the FAED. This framework could be used to assess the trustworthiness of a given FAED score, or the suitability of a feature extraction model for a given domain for related model-based image quality metrics. 

\section*{Acknowledgments}
The project 22HLT05 MAIBAI has received funding from
the European Partnership on Metrology, co-financed from the
European Union’s Horizon Europe Research and Innovation
Programme and by the Participating States. Funding for the
UK partners was provided by Innovate UK under the Horizon
Europe Guarantee Extension.

\bibliographystyle{unsrt}  
\bibliography{references}

\end{document}